\documentclass[10pt, a4paper]{article}
\pdfoutput=1
\usepackage{lrec}
\usepackage{multibib}
\newcites{languageresource}{Language Resources}
\usepackage{graphicx}
\usepackage{tabularx}
\usepackage{soul}

\usepackage{titlesec}
\titleformat{\section}{\normalfont\large\bf\center}{\thesection.}{1em}{}
\titleformat{\subsection}{\normalfont\SmallTitleFont\bf\raggedright}{\thesubsection.}{1em}{}
\titleformat{\subsubsection}{\normalfont\normalsize\bf\raggedright}{\thesubsubsection.}{1em}{}
\renewcommand\thesection{\arabic{section}}
\renewcommand\thesubsection{\thesection.\arabic{subsection}}
\renewcommand\thesubsubsection{\thesubsection.\arabic{subsubsection}}

\usepackage{epstopdf}
\usepackage[utf8]{inputenc}

\usepackage{hyperref}
\usepackage{xstring}

\usepackage{color}
\usepackage[textsize=small]{todonotes}
\usepackage{multirow}

\title{Cross-Lingual Word Embeddings for Turkic Languages}

\name{Elmurod Kuriyozov$^{\ast}$, Yerai Doval$^{\dagger}$, Carlos G\'omez-Rodr\'iguez$^{\ast}$}

\address{ $^{\ast}$Universidade da Coru\~na, CITIC \\
      Grupo LYS, Depto. de Computaci\'on y Tecnologías de la Información\\ Facultade de Inform\'atica, Campus de Elvi\~na, A Coru\~na 15071, Spain\\ 
      \{e.kuriyozov, carlos.gomez\}@udc.es \\ \\
      $^{\dagger}$Grupo COLE, Departamento de Informática E.S. de Enxeñaría Informática, Universidade de Vigo\\ 
       Campus As Lagoas, Ourense 32004, Spain.\\
      yerai.doval@uvigo.es}

\abstract{
There has been an increasing interest in learning cross-lingual word embeddings to transfer knowledge obtained from a resource-rich language, such as English, to lower-resource languages for which annotated data is scarce, such as Turkish, Russian, and many others.
In this paper, we present the first viability study of established techniques to align monolingual embedding spaces for Turkish, Uzbek, Azeri, Kazakh and Kyrgyz, members of the Turkic family which is heavily affected by the low-resource constraint.
Those techniques are known to require little explicit supervision, mainly in the form of bilingual dictionaries, hence being easily adaptable to different domains, including low-resource ones.
We obtain new bilingual dictionaries and new word embeddings for these languages and show the steps for obtaining cross-lingual word embeddings using state-of-the-art techniques. Then, we evaluate the results using the bilingual dictionary induction task. 
Our experiments confirm that the obtained bilingual dictionaries outperform previously-available ones, and that word embeddings from a low-resource language can benefit from resource-rich closely-related languages when they are aligned together. 
Furthermore, evaluation on an extrinsic task (Sentiment analysis on Uzbek) proves that monolingual word embeddings can, although slightly, benefit from cross-lingual alignments.\\ \newline 
 }

\begin{document}

\maketitleabstract

\section{Introduction}

Cross-lingual embeddings are continuous encodings of words or tokens from many languages into the same vector space.
This implies, in principle, that words with similar semantics are represented by similar vectors irrespective of the language they come from.
Previous work has shown that these continuous representations are useful to transfer knowledge from resource-rich languages, mainly English, to many other low-resource languages both in widespread use~\cite{ruder2017survey,doval2019robustness} and also threatened~\cite{adams2017cross}.
An NLP task where knowledge transfer has been of particular interest is machine translation~\cite{zoph2016transfer,gu2018universal}, but there is also work covering many other of the main tasks in the field: PoS tagging~\cite{fang2017model}, dependency parsing~\cite{kulmizev2018multilingual}, or sentiment analysis~\cite{le2016sentiment}.

In this context, to our knowledge, no previous work has covered the case of Turkic languages, a family of languages estimated to be spoken by over 170 million speakers \cite{menz2016speakers} with little to no NLP resources available.
Even though the family comprises dozens of languages, of which at least five have over ten million speakers \cite{menz2016speakers}, 
only Turkish has a reasonable amount of publicly available resources that had been already used in low-resource scenarios.
For the rest of them, we had to resort to extraction from web sites and, in some cases, machine translation to obtain bilingual dictionaries to perform the cross-lingual alignments that we will explain in Section~\ref{sec:methodology}.

In this paper we present, to our knowledge, the first study on the viability of using known techniques for cross-lingual embedding in a realistic setting where most of the languages involved are heavily resource-constrained: in our case, five Turkic languages:  Turkish and Azeri from Oghuz Turkic, Uzbek from Karluk Turkic, as well as  Kazakh and Kyrgyz from Kipchak Turkic language families, while keeping English as the resource-rich counterpart from which we intend to extract the cross-lingual knowledge.
This is in contrast with \cite{adams2017cross} where they artificially lower the amount of data for resource-rich languages.
We are also interested in seeing if the better-equipped Turkish can transfer more specific and potentially useful knowledge to its siblings, complementing that obtained from English.
First, we perform data collection steps, where we obtain new bilingual dictionaries using Google Translate, and we also train monolingual word embeddings from Corpora of Turkic Languages~\cite{baisa2012large} using fastText~\cite{bojanowski2017enriching} for all five above-mentioned Turkic languages.
Then, we perform cross-lingual word embedding alignments using state-of-the-art techniques, namely MUSE~\cite{conneau2018word}, VecMap~\cite{artetxe2018generalizing} and Meemi~\cite{doval2019meemi}. 

Our experiments involve bilingual dictionary induction as an intrinsic task, where we evaluate performances of all five languages. For this task, we first run only a couple of experiments to decide 
which bilingual dictionaries and which monolingual word embeddings perform better, between the already available ones and the ones we newly obtain.
Additionally, we perform sentiment analysis as an extrinsic task to obtain insight on how our cross-lingual word embeddings affect performance on a real task in a monolingual space. This is only done for Uzbek, due to the lack of open data available for all languages we are covering.

Our analyses from obtained results indicate that Meemi~\cite{doval2019meemi} transformations on the VecMap~\cite{artetxe2018generalizing} alignment model give the highest scores for morphologially-rich languages. Apart from that, our figures provide an insight that low-resource languages can benefit more when the reference language is a related one and resource-richer. Furthermore, our sentiment analysis experiment shows that the best aligned word embeddings for Uzbek outperform their initial counterpart.

The resources obtained in the course of the present work are freely available online.\footnote{\url{https://github.com/elmurod1202/crosLingWordEmbTurk}}

\section{Related Work}

Most of the previous work on Turkic languages has been done exclusively on Turkish.
In this case, it has mainly focused on its complex morphology, giving rise to analyzers and disambiguation models~\cite{yuret2006learning,sak2011resources,eryigit2012impact,akyrek2019morphological}.
Nonetheless, other common NLP tasks have been also tackled for Turkish: PoS tagging~\cite{dincer2008suffix,can2016turkish}, dependency parsing~\cite{eryiugit2008dependency,eryiugit2018turkish}, or named entity recognition~\cite{yeniterzi2011exploiting,seker2017extending}.
For a more general overview of the matter, \newcite{oflazer2018turkish} review the state of Turkish NLP.
Although not many, there have been works for other languages rather than Turkish:
for Uzbek~\cite{li2016uzbek,matlatipov2009representation,kuriyozov2019deep}, for Azeri~\cite{fatullayev2008peculiarities,fatullayev2008set,abbasov2010hmm}, for Kazakh~\cite{salimzyanov2013free,sakenovich2016one,yergesh2017sentiment} and for Kyrgyz~\cite{washington2012finite,gormez2011overview}.

Knowledge transfer between languages, and mostly between resource-rich and resource-scarce languages, has drawn much attention in the past years.
This might be attributed to the rise of neural networks as the chosen machine learning technique to tackle all sorts of tasks in NLP: machine translation~\cite{zoph2016transfer,gu2018universal}, PoS tagging~\cite{fang2017model}, dependency parsing~\cite{kulmizev2018multilingual}, or sentiment analysis~\cite{le2016sentiment}.
In general, the internal representations of the linguistic tokens (most of the times, real-valued vectors) obtained by the neural models for the resource-rich language can be considered as representations of the underlying concept, which should be somewhat language-neutral.
Given this assumption, we can then fine-tune the parameters of the network that give rise to these representations with data from the low-resource language to integrate it into the model without needing full retraining.

Since word embeddings follow the same narrative as those internal representations (in fact, they are usually modelled as such), we can think of multilingual, or cross-lingual, embeddings as a vehicle to transfer knowledge between languages.
In this case, we consider the procedure that first obtains monolingual embeddings and then aligns them into a shared space.
Starting from the work of \newcite{mikolov2013exploiting}, this approach has seen many contributions~\cite{faruqui2014improving,lazaridou2015hubness,zhang2016ten,smith2017offline}.
Notable methods recently developed are VecMap~\cite{artetxe2018generalizing}, MUSE~\cite{conneau2018word}, and then Meemi~\cite{doval2019meemi} as a way to further improve the integration of the cross-lingual space.
Regarding low-resource languages, some analysis papers such as~\cite{ruder2017survey,sogaard2018limitations,doval2019robustness,glavas2019evaluate} study the viability of these methods in adverse scenarios, and others even resort to endangered languages, such as \cite{adams2017cross}.


\section{Methodology}
\label{sec:methodology}

\subsection{Cross-lingual embeddings}

We start by building monolingual embedding models for each of the $n$ languages considered. This could be done through any of the well-known word embedding algorithms; e.g., word2vec~\cite{Mikolov2013}, GloVe~\cite{pennington2014glove}, or fastText~\cite{bojanowski2017enriching}, although we will use the latter in this work.
Since these models produce isolated vector spaces, we then apply a pairwise alignment step that maps $n-1$ of the models into a reference one, which corresponds to English language in our case, that remains fixed throughout the process.
The most widely-used algorithms for this, such as VecMap~\cite{artetxe2018generalizing} and MUSE~\cite{conneau2018word}, rely on orthogonal transformations learned on bilingual dictionaries that preserve the internal structure of the monolingual spaces while they become integrated into the same cross-lingual space (that of the reference language).

Although several authors show that maintaining the internal monolingual structures avoids overfitting to the cross-lingual objective, which would degrade the monolingual performance~\cite{artetxe2016learning}, \newcite{doval2019meemi} have found that applying a simple post-processing step that modifies those structures improves the cross-lingual performance of the models without compromising on the monolingual side; quite the contrary, in fact, as this step improves the monolingual performance.
This post-processing step involves taking already aligned vector spaces using alignment models that keep the internal structure of the monolingual spaces, then applies an additional transformation that map the vector representations of both word and its translation onto their average, thereby creating a cross-lingual vector space which intuitively corresponds to the average of the two aligned monolingual vector spaces. The mapping introduced in this step also uses the same bilingual lexicon that was used for initial alignment.

This is of crucial importance to us since it shows that transferring knowledge between languages is not only possible but potentially beneficial.

\subsection{Data Collection}
To obtain cross-lingual word embeddings, we first need monolingual word embeddings for both source and all target languages, as well as dictionaries from the source language to each of the target languages.

It is worth noting that some of our chosen five Turkic languages currently use two or
three writing systems. For our embeddings, we decided to focus on the scripts that are prevailing in recent texts: Latin for Turkish, Uzbek and Azerbaijani, and Cyrillic (with extensions) for Kazakh and Kyrgyz.

\subsubsection{Obtaining Word Embeddings}
First, we obtained preexisting fastText pre-trained word embeddings for our target languages~\cite{grave2018learning}. Apart from that, considering the fact that all target languages are low-resource and there is always room for improvement, we also decided to train our own word embeddings and use them for experiments. For that purpose, we obtained a collection of large corpora for Turkic languages~\cite{baisa2012large} and trained new fastText embeddings.
Size statistics for our obtained word embeddings are shown in Table~\ref{table:embedding-sizes}.

\begin{table}
\begin{center}
\begin{tabular}{lccc}
\hline
\textbf{Language} & \textbf{Tokens} & \textbf{Types} & \textbf{Embeddings} \\ 
\hline
Turkish & 3,370M & 20.5M & 6M \\
Uzbek   & 18M    & 626K  & 200K  \\
Azeri   & 92M    & 1.7M  & 554K  \\
Kazakh  & 136M   & 2.4M  & 803K  \\
Kyrgyz  & 19M    & 684K  & 228K  \\
\hline
\end{tabular}
\caption{Vocabulary sizes of obtained fastText word embeddings. 
The Tokens column represents the number of word occurrences in each language corpus, whereas Types shows the number of unique words and the Embeddings column represents the number of vectors in the trained fastText word embeddings. The number of vectors is smaller than the number of unique words because infrequent words do not get a vector.}
\label{table:embedding-sizes}
 \end{center}
\end{table}
We shall be using both embeddings in our experiments to evaluate their performance.

\subsubsection{Obtaining Dictionaries}

\label{sec:dictionaries}

Proper dictionaries that are both free to use and adequate in terms of size and accuracy are not always available for low-resource languages. 
Thus, while we tried to use pre-existing resources where available, for some languages we had to resort to generating ad-hoc dictionaries using a translation API. In particular, we obtained dictionaries in the following two ways:
\begin{enumerate}

\item{From available pre-existing dictionaries.} 
We first tried to find required dictionaries for our experiments. We were able to obtain them only for three languages: Turkish, Uzbek and Kazakh. For Turkish-English, we found a  bilingual dictionary available at MUSE~\cite{conneau2018word}. For Uzbek, we extracted an Uzbek-English dictionary from The Uzbek Glossary.\footnote{\url{http://www.uzbek-glossary.com}} Similarly, for Kazakh language, we extracted a dictionary from The Leneshmid Dictionary website.\footnote{\url{http://kazakh-glossary.com/table1list.php}} Dictionaries for other languages were not found due to either none being available (to our knowledge), not being free to use or difficulty to extract a sufficient amount of words. The sizes of the dictionaries obtained in this way are shown in   Table~\ref{table:web-dict-sizes}.

\begin{table}
\begin{center}
\begin{tabular}{lc}
\hline
\textbf{Language} & \textbf{Dictionary Size} \\ 
\hline
Turkish - English & 68306 \\
Uzbek - English & 4042  \\
Kazakh - English & 15200 \\
\hline
\end{tabular}
\caption{Number of words in obtained dictionaries}
\label{table:web-dict-sizes}
 \end{center}
\end{table}

\item{Using Google Translate.} In order to fill the gap to provide bilingual dictionaries for the rest of our target languages, as well as trying to get better ones for the ones where a dictionary was already available, we used Google Translate.
To do so, we first obtained a list of the most frequent 30K English words from SketchEngine,\footnote{\url{https://app.sketchengine.eu} requires user authorization to access language corpora and resources.} and translated them into all target languages using 
Google Translate.
To provide high translation accuracy, the next step was to reverse-translate resulting translations from target languages back to English, 
and keeping only the ones that are translated back to the initial English word. This step decreased the size of resulting dictionaries to about a third.
Thus, to further increase the coverage of the dictionaries without compromising quality, and in particular to ensure that frequent words were covered, we then
merged the results with bilingual dictionaries obtained from the 1000 Most Common Words website.\footnote{\url{https://1000mostcommonwords.com}} Then, a cleaning process was undertaken by removing duplicate lines and dictionary translations containing more than one token (word). The last step was to split the resulting dictionaries into training and test sets, by randomly choosing 500 words from each language for the test sets, and using the rest as training sets. Table~\ref{table:gtrans-dicts-sizes} reports statistics about the size of the dictionaries obtained in this way.
\end{enumerate}

\begin{table*}
\begin{center}
\begin{tabular}{lccccc}
\hline
\textbf{Language} & \textbf{At First} & \textbf{After Reverse Translation} & \textbf{After Cleaning} & \textbf{Test set} & \textbf{Training  \ set} \\ 
\hline
Turkish & \multirow{5}{*}{30000} & 13186 & 9850 & \multirow{5}{*}{500}   & 9350 \\
Uzbek   &  & 12746  & 8458 &  & 7958 \\
Kazakh  &  & 14125  & 8954 &  & 8454 \\
Azeri   &  & 11113  & 7922 &  & 7422 \\
Kyrgyz  &  & 14907  & 8474 &  & 7974 \\  
\hline
\end{tabular}
\caption{Number of words during the process of obtaining dictionaries from Google Translate. The indicated numbers in second column (At First) and the second to last(Test set) apply to all rows, meaning that for all languages it starts with initial 30000 words and the resulting clean dictionaries are split in a way that 500 word pairs are given for test set, remaining for training set.}
\label{table:gtrans-dicts-sizes}
 \end{center}
\end{table*}

\section{Experiments and Results}
\label{section:exp&res}

Our experiments involve both intrinsic and extrinsic evaluation of our obtained cross-lingual embeddings. As an intrinsic task, we perform bilingual dictionary induction for the five languages we are covering in this paper. As an extrinsic task, we use sentiment analysis for Uzbek language only, due to the lack of open data available for other languages. This provides insight about how our cross-lingual word embeddings affect performance on a real task in a monolingual space.

Cross-lingual embeddings used for both experiments were trained under the following conditions: 
\begin{itemize}
    \item \textbf{Monolingual word embeddings} were obtained from available pre-trained word vectors ~\cite{grave2018learning} trained on CommonCrawl\footnote{\url{https://commoncrawl.org/}} and Wikipedia\footnote{\url{https://en.wikipedia.org}} using fastText~\cite{bojanowski2017enriching}. A second set of embeddings was also obtained by training on the Corpora of Turkic Languages from ~\cite{baisa2012large} using fastText with its default hyper-parameters, except for the minimal number of word occurrences (minCount) which was set to 3, due to the fact that all these Turkic languages are highly agglutinative,\footnote{\url{https://en.wikipedia.org/wiki/Turkic_languages}} so there is a high variety of word forms that can occur very few times. Both monolingual embeddings have a vector dimension of 300. 

    \item \textbf{Bilingual dictionaries.} 
    As mentioned in Section \ref{sec:dictionaries}, we used already available dictionaries obtained from websites in the case of English to Turkish, Uzbek and Kazakh; and we also obtained our own dictionaries using Google Translate from English to all the Turkic languages covered in this paper.
    Table~\ref{table:web-dict-sizes} and Table~\ref{table:gtrans-dicts-sizes} report the sizes of 
    each of these sets of dictionaries.
    
    \item \textbf{Cross-lingual alignment.} By default, in order to obtain alignments for monolingual embeddings, we considered English as a source language and all these five Turkic languages as target ones. In some of the experiments, we also tried models where some of the Turkic languages are added as additional source languages with the rest remaining as target languages. We compared the results of alignments using the three models' open-source implementations: MUSE~\cite{conneau2018word} and both the orthogonal and multistep versions of VecMap~\cite{artetxe2018generalizing}, as well as Meemi~\cite{doval2019meemi} transformations based on two previously mentioned models. They were all used with the recommended parameters. 
\end{itemize}

\subsection{Bilingual Dictionary Induction.} 

The dictionary induction task considers the problem of retrieving the translation of a given input (i.e. list of words from the source language) to a target language using word embeddings aligned in a same space. For this task we used three alignments: fastText word embeddings aligned using bilingual dictionaries obtained from websites for English-Turkish, English-Uzbek and English-Kazakh, fastText word embeddings aligned using dictionaries obtained from Google Translate, and embeddings obtained from Turkic corpora aligned using dictionaries obtained from Google Translate. The latter two alignments are for all five Turkic languages we are covering in this paper.
During experiments, we first run only a couple of cross-lingual alignment techniques in order to decide what dictionaries and word embeddings to use for better performance, and then compare all alignment techniques on the chosen ones.

Test sets were made of dictionaries with 500 words for each of the languages, obtained as a split from the dictionaries generated from Google Translate. In order to make all results comparable, we used the same test sets over all experiments and also made sure that no test set overlapped with the training set of the respective language. 

The task takes an input word from the source language and outputs 
a ranked list of translations to the target language.
This is implemented as a simple nearest neighbour search using cosine distance.
The performance of the embeddings alignment is evaluated using the precision at \textit{k} (\textit{P@k}) metric. This metric 
counts the percentage of correct answers (translation pairs) that are among the top $k$ ranked candidates.


\textbf{Results}. 
Here we discuss the results of our experiments, for the different combinations of embeddings and bilingual dictionaries:
\begin{itemize}
    \item \textbf{FastText embeddings} aligned using \textbf{available dictionaries}. 
    Table~\ref{table:results-web-dict} shows the dictionary induction task results of fastText available word embeddings aligned using dictionaries obtained from websites for Turkish, Uzbek and Kazakh. The other languages were not covered in this experiment due to the lack of available dictionaries. The results indicate that the Meemi transformation based on the orthogonal VecMap model outperforms VecMap's baseline model for all three languages. Dictionary induction on English-Turkish is considerably better than the other two cases, as more than 50\% of words are retrieved at the first prediction (P@1), and in roughly 80\% cases the correct translation is among 10 nearest words (P@10). For other language pairs we obtain worse performance, especially for English-Uzbek, mostly due to the relatively small size of the dictionary trained, as well as the fact that word embeddings were not trained on large corpora. Another reason why the English-Uzbek alignment obtained the worst results in this task is because the dictionary obtained from websites was in all lowercase letters, which resulted in many out-of-vocabulary (OOV) words during mapping.

    \begin{table*}
    \begin{center}
    \begin{tabular}{l|ccc|ccc|ccc}
    \hline
    \textbf{Language} & \multicolumn{3}{c|}{\textbf{Turkish}} & \multicolumn{3}{c|}{\textbf{Uzbek}} & \multicolumn{3}{c}{\textbf{Kazakh}} \\
    \hline
    \textbf{Prediction}    & \textbf{P@1}   & \textbf{P@5}   & \textbf{P@10}  & \textbf{P@1}   & \textbf{P@5}   & \textbf{P@10}  & \textbf{P@1}   & \textbf{P@5}   & \textbf{P@10} \\ 
    \hline
    \textbf{VecMap}  & 53.3   & 72.4   & 76.8    & 4.6    & 12.8   & 18.0    & 28.5   & 49.5   & 54.7    \\
    \textbf{Meemi(VecMap)} & 53.9   & 75.3   & 78.4    & 4.0    & 15.7   & 19.7    & 33.6   & 51.5   & 60.8   \\
    \hline
    \end{tabular}
    \caption{Results of Bilingual Dictionary Induction analysis. FastText word embeddings and dictionaries obtained from websites were used for experiments.  \textbf{VecMap} is the orthogonal VecMap model, \textbf{Meemi(VecMap)} is Meemi transformations on the orthogonal VecMap model. \textbf{P@1}, \textbf{P@5} and \textbf{P@10} metrics indicate the percentages of correct answers(translation pairs) that are among the top ranked 1,5 and 10 candidates respectively.}
    \label{table:results-web-dict}
    \end{center}
    \end{table*} 

    \item \textbf{New embeddings}  aligned using \textbf{new dictionaries}. 
    As can be seen in Table~\ref{table:results-new-emb}, dictionary induction results for English-Turkish and English-Azeri 
    are better than for other languages,
    mainly due to the larger size and better quality of both dictionary and word embeddings, while results for other languages are quite related to the size of corpora their embeddings were trained with. It is worth mentioning that there is a considerable improvement in the results for English-Uzbek when compared to the previous results, with slightly more than 30\%  P@1 and 50\% P@5, proving that our obtained dictionaries not only are larger, but also have good quality. Even though the dictionaries they were all trained with have similar sizes, induction results of English-Kyrgyz have the smallest scores, mainly due to the fact that the word embeddings were trained on a very small corpus.
    When comparing the different cross-lingual embedding techniques, Meemi transformations on orthogonal VecMap model outperform the orthogonal VecMap baseline model again for all five languages in all metrics.

    \begin{table*}
    \begin{small}
    \begin{tabular}{l|ccc|ccc|ccc|ccc|ccc}
    \hline
    \textbf{Language} & \multicolumn{3}{c|}{\textbf{Turkish}} & \multicolumn{3}{c|}{\textbf{Uzbek}} & \multicolumn{3}{c|}{\textbf{Azeri}} & \multicolumn{3}{c|}{\textbf{Kazakh}} & \multicolumn{3}{c}{\textbf{Kyrgyz}} \\
    \hline
    \textbf{Prediction}    &  \textbf{@1}   &  \textbf{@5}   & \textbf{@10}    &  \textbf{@1}   &  \textbf{@5}   & \textbf{@10}     &  \textbf{@1}   &  \textbf{@5}   & \textbf{@10}     &  \textbf{@1}   &  \textbf{@5}   & \textbf{@10}     &  \textbf{@1}   &  \textbf{@5}   & \textbf{@10} \\ 
    \hline
    \textbf{VecMap} & 50.1  & 70.2  & 74.3   & 24.8  & 42.2  & 47.7   & 37.5  & 58.7  & 67.8   & 32.3  & 52.3  & 59.7   & 16.4  & 30.7  & 35.7   \\
    \textbf{Meemi(VecMap)} & 52.5  & 72.4  & 77.9   & 31.5  & 50.5  & 54.7   & 43.9  & 67.3  & 73.3   & 38.9  & 58.7  & 65.3   & 21.7  & 35.7  & 40.2  \\
    \hline
    \end{tabular}
    \caption{Results of Bilingual Dictionary Induction analysis. Newly obtained word embeddings from Turkic Languages Corpora trained using fastText and dictionaries obtained from Google Translate were used for experiments.  \textbf{VecMap} is the orthogonal VecMap model, \textbf{Meemi(VecMap)} is Meemi transformations on the orthogonal VecMap model.} 
    \label{table:results-new-emb}
    \end{small}
    \end{table*}

    \item \textbf{FastText embeddings} aligned using \textbf{new dictionaries}.
    Table~\ref{table:results-big} shows the results of the bilingual dictionary induction task obtained from alignments trained on fastText word embeddings and dictionaries obtained from Google Translate. Corresponding to the results from Table~\ref{table:results-new-emb}, the task is performed better for Turkish and Azeri, while the results for Kazakh and Uzbek lag behind, and Kyrgyz has the worst scores. As it turns out, this alignment outperforms previous alignments for all languages, except for Uzbek where it has slightly lower scores than the scores shown in Table~\ref{table:results-new-emb}. Since this setting obtained the best results for most languages, we chose it for our main comparison between cross-lingual alignment models, by performing more alignments using a larger variety of techniques and more than one source language option. The alignment models included in this experiment are: \textbf{MUSE}\footnote{\url{https://github.com/facebookresearch/MUSE}} from \newcite{conneau2018word};  \textbf{VecMap} (orthogonal VecMap model) and \textbf{MVecMap} (multistep VecMap)\footnote{\url{https://github.com/artetxem/vecmap}} from \newcite{artetxe2018generalizing}; \textbf{Meemi(MUSE)} and \textbf{Meemi(VecMap)}\footnote{\url{https://github.com/yeraidm/meemi}} from  \newcite{doval2019meemi}, all of these models use English as the source language. \textbf{En-Tr-*} corresponds to a Meemi(Vecmap) alignment with two source languages: English and Turkish, where the rest are target languages; and \textbf{En-Tr-Az-*} is a Meemi(Vecmap) alignment with three source languages: English, Turkish and Azeri, where rest are target languages. Finally, \textbf{All(Meemi)} includes all English and five Turkic languages aligned in a same space using Meemi(Vecmap). 
    
    By analysing Table~\ref{table:results-big}, one can see similar trends as previous alignment results where Turkish has the best scores, followed by Azeri, whilst Kyrgyz has the lowest ones. From the comparison over alignment models, it can be concluded that the orthogonal VecMap model outperforms the MUSE model for all languages and metrics, while MVecMap, which is an extra transformation step done on the VecMap model, outperforms both of its predecessors and Meemi(MUSE) - Meemi transformations on MUSE alignment for all languages. Meemi transformations on the orthogonal VecMap model (Meemi(VecMap)) outperform all previously mentioned alignments most of the time, proving itself the best model to continue with more experiments where more than  one source language are used.
    
    By choosing Meemi(VecMap) as the best model, we widened the experiment by adding one or more relatively resource-rich languages as source ones alongside English, to see if the low-resource languages would benefit from this alignment. Turkish and Azeri are both resource-richer and closely related to the target languages. Dictionary induction scores show that all the target languages benefit from this alignment. When Turkish was added alongside English as a source language, Azeri and Kyrgyz benefited considerably from this alignment, when Azeri was also added as the third source language, Uzbek and Kazakh scores increased, indicating that the more closely related is the language that is added to the alignment, the more the low-resource language benefits from that alignment.
    
    Lastly, we tried to align all five languages alongside English and obtained scores of bilingual dictionary induction. Results show that resource-richer languages (Turkish and Azeri) hugely decrease in performance while the lowest-resource one (Kyrgyz) reaches its best score in the P@1 metric.

\begin{table*}[ht]
\begin{small}
\begin{tabular}{l|ccc|ccc|ccc|ccc|ccc}
\hline
\textbf{Language} & \multicolumn{3}{c|}{\textbf{Turkish}} & \multicolumn{3}{c|}{\textbf{Uzbek}} & \multicolumn{3}{c|}{\textbf{Azeri}} & \multicolumn{3}{c|}{\textbf{Kazakh}} & \multicolumn{3}{c}{\textbf{Kyrgyz}} \\
\hline
 

\textbf{Prediction}    &  \textbf{@1}   &  \textbf{@5}   & \textbf{@10}    &  \textbf{@1}   &  \textbf{@5}   & \textbf{@10}     &  \textbf{@1}   &  \textbf{@5}   & \textbf{@10}     &  \textbf{@1}   &  \textbf{@5}   & \textbf{@10}     &  \textbf{@1}   &  \textbf{@5}   & \textbf{@10} \\ 
\hline

\textbf{MUSE} & 51.7 & 70.0 & 75.9 & 20.4 & 38.2 & 45.4 & 39.4 & 56.0 & 62.8 & 36.4 & 53.6 & 58.5 & 14.5 & 30.4 & 36.3 \\
\textbf{Meemi(MUSE)} & 58.0 & 75.1 & 78.4 & 23.9 & 41.5 & 49.4 & 47.4 & 64.1 & 68.1 & 41.0 & 59.9 & 66.4 & 22.7 & 35.4 & 42.2 \\
\textbf{VecMap} & 55.4 & 75.1 & 79.4 & 23.7 & 40.2 & 46.4 & 45.8 & 43.0 & 70.5 & 38.0 & 59.0 & 66.4 & 17.7 & 36.9 & 41.3 \\
\textbf{MVecMap} & 58.4 & 76.3 & 80.3 & 28.2 & 47.0 & 52.4 & 48.9 & 66.7 & 73.6 & 43.6 & 60.8 & 67.8 & 23.9 & 37.8 & 45.7 \\
\textbf{Meemi(VecMap)} & \textbf{59.1} & \textbf{76.8} & \textbf{81.1} & \textbf{28.2} & 44.7 & 51.9 & 47.8 & 67.2 & 74.2 & 43.8 & 62.0 & \textbf{70.6} & 23.3 & 40.4 & 46.9 \\
\textbf{En-Tr-*} & x & x & x & 26.5 & 46.2 & 53.3 & \textbf{51.3} & \textbf{70.9} & \textbf{75.3} & 43.4 & \textbf{63.2} & 70.2 & 23.6 & \textbf{41.9} & \textbf{48.1} \\
\textbf{En-Tr-Az-*} & x & x & x & 27.9 & \textbf{47.3} & \textbf{53.6} & x & x & x & \textbf{44.3} & 62.5 & 69.5 & 22.1 & 41.0 & 47.8 \\
\textbf{All(Meemi)}    & 49.4 & 73.7 & 77.6 & 25.1 & 43.3 & 46.2 & 44.7 & 62.6 & 69.2 & 39.9 & 57.6 & 61.5 & \textbf{25.4} & 41.6 & 47.2 \\   

\hline
\end{tabular}
\caption{Results of Bilingual Dictionary Induction analysis using different alignment models. FastText word embeddings and dictionaries obtained from Google Translate were used for experiments of all languages. \textbf{MUSE} is MUSE model, \textbf{Meemi(MUSE)} is Meemi transormations on MUSE results, \textbf{VecMap} is orthogonal VecMap model, \textbf{MVecMap} is multistep VecMap model, \textbf{Meemi(VecMap)} is Meemi transformations on orthogonal VecMap model, \textbf{En-Tr-*} is Tri-Lingual alignment of Meemi transformations on VecMap where English and Turkish play as a source language, \textbf{En-Tr-Az-*} stands for results of 4-language alignment of Meemi transformations on VecMap where English-Turkish-Azeri languages play as a source and the rest as a target languages, \textbf{All(Meemi)} represents the results of all six languages aligned in the same space by Meemi transformations on VecMap. Highest scores are highlighted with bold faces. \textbf{x} indicates that the task is not applicable because the language plays the role of a source together with English.}
\label{table:results-big}
\end{small}
\end{table*}

\end{itemize}

\subsection{Evaluation on a Monolingual Space}

In order to check if there is any improvement on the monolingual vector space of a low-resource language after performing alignment with other language spaces (preferably related languages and resource-rich ones), we ran a small experiment on Uzbek,
as a sentiment analysis dataset has recently been presented for this language, together with experiments using various machine learning and deep learning architectures to implement sentiment analysis~\cite{kuriyozov2019deep}.
One of the models presented on said paper used recurrent neural networks (RNNs) trained with fastText word embeddings. We obtained the shared code and datasets and ran the same model, this time replacing fastText word embeddings with the best vector space obtained by the Meemi model trained on orthogonal VecMap using Uzbek fastText word embeddings aligned with English-Turkish-Azeri languages by Google Translate dictionaries. Both datasets presented by \cite{kuriyozov2019deep} (a ``MANUAL'' dataset of 4.3K manually annotated reviews and a ``TRANSLATED'' dataset of 20K reviews translated from English to Uzbek) were used for experiments.

\paragraph{Experimental Settings.} To run sentiment analysis experiments, the same settings from ~\cite{kuriyozov2019deep} were used.  A bidirectional network of 100 GRUs was used, and the output of the hidden layer is the concatenation of the average and max pooling of the hidden states. The final output is obtained by a sigmoid activation function applied on the previous layer. As in the original paper, the Adam optimization algorithm with standard parameters (learning rate $\alpha=0.0001$, exponential decay rates $\beta_1 =0.9$, $\beta_2 =0.999$) was chosen for training.
Binary cross-entropy was used as loss function. The training/test split was preserved.

\begin{table}
\begin{center}
\begin{tabular}{lcc}
\hline
\multicolumn{1}{c}{\textbf{Dataset}} & \textbf{FastText Score*} & \textbf{Our Score} \\ \hline
Manual Dataset   & 0.8782   & \textbf{0.8825}    \\
Translated Dataset & 0.8832   & \textbf{0.8876}   
\end{tabular}
\caption{Sentiment analysis results on RNN model pre-trained with aligned Uzbek word embeddings. 
Best results are highlighted in bold.
*FastText scores shown in the table are directly taken from the original paper.}
\label{table:sentanalysis}
 \end{center}
\end{table}
 
\paragraph{Results.} Table~\ref{table:sentanalysis} shows the results of analyses from both the original paper and our experiments. The RNN model trained with aligned word embeddings slightly outperforms the same model trained with fastText word embeddings in both the Manual and Translated datasets. This shows that a low-resource language can benefit from the Meemi model when aligned with resource-rich and/or similar languages. 

\section{Conclusion and Future Work}

In this paper we obtained new bilingual dictionaries for five Turkic languages: Turkish, Uzbek, Azeri, Kazakh, Kyrgyz, using the Google Translate API. We also obtained better (compared to what's available so far) word embeddings for the Uzbek language, trained from Corpora of Turkic languages~\cite{baisa2012large} using fastText~\cite{bojanowski2017enriching}. 

We also analysed cross-lingual word embeddings for all five Turkic languages mentioned above using state-of-the-art alignment methods, namely MUSE~\cite{conneau2018word}, VecMap~\cite{artetxe2018generalizing} and Meemi~\cite{doval2019meemi}. 
We tested our alignments with a bilingual dictionary induction task, where we presented the scores for all five languages for the different dictionaries and word embeddings used for alignment of vector spaces.

Moreover, this task gave us an opportunity of analysing the quality of existing and newly obtained bilingual dictionaries as well as word embeddings, which were used for embedding alignments. 

We witnessed a quality improvement on a monolingual vector space after cross-lingual alignment by running an additional task: sentiment analysis on Uzbek language, using all the datasets and methods from \cite{kuriyozov2019deep}.

This work proves itself to be, to our knowledge, the first one of its kind for Turkic languages and the results obtained from tasks still have room for improvement.

Future work could include, first of all, improving fundamental resources like monolingual word embeddings and bilingual dictionaries for lower-resourced languages to improve the results of cross-lingual tasks. Secondly, while we only covered five Turkic languages, the language scope can be widened by including more languages in this family, which will be convenient for intra-language-family NLP tasks, like for example machine translation among Turkic languages. 
Lastly, following previous work and also for the sake of easiness in obtaining bilingual dictionaries, we chose English as the reference language. This way of aligning languages using English, which is not related to Turkic languages, could be improved by choosing a related language, most probably a resource-rich one (like Turkish for the Turkic family), as a reference language instead. It would be interesting to see if this could produce improvements in the results.

\section{Acknowledgements}
Elmurod Kuriyozov received funding from the  ANSWER-ASAP project (TIN2017-85160-C2-1-R) from MINECO, and from Xunta de Galicia (ED431B 2017/01, ED431G 2019/01). He is also funded for his PhD by El-Yurt-Umidi Foundation under the Cabinet of Ministers of the Republic of Uzbekistan.

Yerai Doval has been supported by the Spanish Ministry of Economy, Industry and Competitiveness (MINECO) through the ANSWER-ASAP project (TIN2017-85160-C2-2-R); by the Spanish State Secretariat for Research, Development and Innovation (which belongs to MINECO) and the European Social Fund (ESF) through a FPI fellowship (BES-2015-073768) associated to TELEPARES project (FFI2014-51978-C2-1-R); and by the Xunta de Galicia through TELGALICIA research network (ED431D 2017/12).

Carlos Gómez-Rodríguez has received funding from the European Research Council (ERC), under the European Union’s Horizon 2020 research and innovation programme (FASTPARSE, grant agreement No 714150), from the ANSWER-ASAP project (TIN2017-85160-C2-1-R) from MINECO, and from Xunta de Galicia (ED431B 2017/01, ED431G 2019/01, and Oportunius program).







\section{Bibliographical References}\label{reference}

\bibliographystyle{lrec}
\bibliography{lrec2020W-xample-kc}


\end{document}